\documentclass{frontiers}

\usepackage{url}
\usepackage{amsmath}
\usepackage{latexsym}
\usepackage{mathtools}
\usepackage{graphicx}
\usepackage{caption}
\usepackage{subfig}
\usepackage{setspace}
\usepackage{courier}
\usepackage{algorithm}
\usepackage{algorithmic}
\usepackage{textcomp}
\usepackage{units}
\usepackage{nicefrac}
\usepackage{epstopdf}
\usepackage{color}
\usepackage{ulem}
\usepackage[english]{babel}

\DeclarePairedDelimiter\norm{\lVert}{\rVert}%

\copyrightyear{}
\pubyear{}
\def\journal{Neurosciences}
\def\DOI{}
\def\articleType{}

\def\firstAuthorLast{Zhong {et~al}} 
\def\Authors{Junpei Zhong\,$^{1,*}$, Angelo Cangelosi\,$^{2}$ and Stefan Wermter\,$^1$}
\def\Address{$^{1}$Department of Computer Science, University of Hamburg, Hamburg, Germany \\
$^{2}$School of Computing and Mathematics, University of Plymouth, Plymouth, UK}
\def\corrAuthor{Junpei Zhong}
\def\corrAddress{University of Hamburg, \\
Department of Computer Science, Knowledge Technology, \\
 Vogt-K\"{o}lln-Stra\ss e 30,\\
 22527 Hamburg, Germany}
\def\corrEmail{zhong@informatik.uni-hamburg.de}

\begin{document}
\onecolumn
\firstpage{1}

\title[A self-organizing pre-symbolic neural model]{Towards a self-organizing pre-symbolic neural model representing sensorimotor primitives}
\author[\firstAuthorLast ]{\Authors}
\address{}
\correspondance{}
\editor{}
\topic{Research Topic}

\maketitle
\begin{abstract}
The acquisition of symbolic and linguistic representations of sensorimotor behavior is a cognitive process performed by an agent when it is executing and/or observing own and others' actions. According to Piaget's theory of cognitive development, these representations develop during the sensorimotor stage and the pre-operational stage.  We propose a model that relates the conceptualization of the higher-level information from visual stimuli to the development of ventral/dorsal visual streams. This model employs neural network architecture incorporating a predictive sensory module based on an RNNPB (Recurrent Neural Network with Parametric Biases) and a horizontal product model. We exemplify this model through a robot passively observing an object to learn its features and movements. During the learning process of observing sensorimotor primitives, i.e. observing a set of trajectories of arm movements  and its oriented object features, the pre-symbolic representation is self-organized in the parametric units.  These representational units act as bifurcation parameters, guiding the robot to recognize and predict various learned sensorimotor primitives. The pre-symbolic representation also accounts for the learning of sensorimotor primitives in a latent learning context.
\vspace{1em}

\tiny
  \section{Keywords:} Pre-symbolic Communication, Sensorimotor Integration, Recurrent Neural Networks, Parametric Biases, Horizontal Product   
\end{abstract}
\section{Introduction}
Although infants are not supposed to acquire the symbolic representational system at the sensorimotor stage, based on Piaget's definition of infant development, the preparation of language development, such as a pre-symbolic representation for conceptualization, has been set at the time when the infant starts babbling (\cite{mandler1999preverbal}).
Experiments have shown that infants have established the concept of animate and inanimate objects, even if they have not yet seen the objects before (\cite{gelman1981development}). Similar phenomena also include the conceptualization of object affordances such as the conceptualization of containment (\cite{bonniec1985visual}). This conceptualization mechanism is developed at the sensorimotor stage to represent sensorimotor primitives and other object-affordance related properties. 

During an infants' development at the sensorimotor stage, one way to learn affordances is to interact with objects using tactile perception, observe the object from visual perception and thus learn the causality relation between the visual features, affordance and movements as well as to conceptualize them. This learning starts with the basic ability to move an arm towards the visual-fixated objects in new-born infants (\cite{von1982eye}), continues through object-directed reaching at the age of 4 months (\cite{streri1993seeing,corbetta2009seeing}), and can also be found during the object exploration of older infants (c.f.~\cite{mandler1992foundations, ruff1984infants}). From these interactions leading to visual and tactile percepts, infants gain experience through the instantiated `bottom-up' knowledge about object affordances and sensorimotor primitives. Building on this, infants at the age of around 8-12 months gradually expand the concept of object features, affordances and the possible causal movements in the sensorimotor context (\cite{gibson1988exploratory, newman2001development, rocha2006impact}).
For instance, they realize that it is possible to pull a string that is tied to a toy car to fetch it instead of crawling towards it.
{An associative rule has also been built that connects conceptualized visual feature inputs, object affordance and the corresponding frequent auditory inputs of words, across various contexts (\cite{romberg2010statistical}).
At this stage, categories of object features are particularly learned in different contexts due to their affordance-invariance (\cite{bloom1993words}). }

Therefore the integrated learning process of the object's features, movements according to the affordances, and other knowledge is a globally conceptualized  process through visual and tactile perception. 
This conceptualized learning is a precursor of a pre-symbolic representation of language development.
{This learning is the process to form an abstract and simplified representation for information exchange and sharing\footnote{For comparison of conceptualization between engineering and language perspectives, see \cite{gruber1994ontology,bowerman2001language}.}. 
To conceptualize from visual perception, it usually includes a planning process: first the speaker receives and segments visual knowledge in the perceptual flow into a number of states on the basis of different criteria, then the speaker selects essential elements, such as the units to be verbalized, and last the speaker constructs certain temporal perspectives when the events have to be anchored and linked (c.f. \cite{habel1999processes, von2003processes}). 
Assuming this planning process is distributed between ventral and dorsal streams, the conceptualization process should also emerge from the visual information that is perceived in each stream, } {associating the distributed information in both streams. 
As a result, the candidate concepts of visual information are statistically associated with the input stimuli. For instance, they may represent a particular visual feature with a particular class of label (e.g. a particular visual stimuli with an auditory wording `circle') (\cite{chemla2009categorizing}). Furthermore, the establishment of such links also strengthens the high-order associations that generate predictions and generalize to novel visual stimuli (\cite{yu2008statistical}).
Once the infants have learned a sufficient number of words, they begin to detect a particular conceptualized cue with a specific kind of wording. At this stage, infants begin to use their own conceptualized visual `database' of known words to identify a novel meaning class and possibly to extend their wording vocabulary (\cite{smith2002object}).  
Thus, this associative learning process enables the acquisition and the extension of the concepts of domain-specific information (e.g. features and movements in our experiments) with the visual stimuli. }

This conceptualization will further result in a pre-symbolic way for infants to communicate when they encounter a conceptualized object and intend to execute a correspondingly conceptualized well-practised sensorimotor action towards that object. For example, behavioral studies showed that when 8-to-11-month-old infants are unable to reach and pick up an empty cup, they may point it out to the parents and execute an arm movement intending to bring it to their lips. The conceptualized shape of a cup reminds infants of its affordance and thus they can communicate in a pre-symbolic way. Thus, the emergence from the conceptualized visual stimuli to the pre-symbolic communication also gives  further rise to the different periods of learning nouns and verbs in infancy development (c.f. \cite{gentner1982nouns, tardif1996nouns, bassano2000early}). 
This evidence supports that the production of verbs and nouns are not correlated to the same modality in sensory perception: experiments performed by \cite{kersten1998examination} suggest that nouns are more related to the movement orientation caused by the intrinsic properties of an object, while verbs are more related to the trajectories of an object.
Thus we argue that such differences of acquisitions in lexical classes also relate to the conceptualized visual ventral and dorsal streams. The finding is consistent with~\cite{damasio1993nouns}'s hypothesis that verb generation is modulated by the perception of conceptualization of movement and its spatio-temporal relationship. 

For this reason, we propose that the conceptualized visual information, which is a prerequisite for the pre-symbolic communication, is also modulated by perception in two visual streams. 
{Although there  have been studies of modeling the functional modularity in the development of ventral and dorsal streams (e.g. \cite{jacobs1991task, mareschal1999computational}), the bilinear models of visual routing (e.g. \cite{Olshausen93,bergmann2011self,memisevic2007unsupervised}), in which a set of control neurons dynamically modifies the weights of the `what' pathway on a short time scale, or transform-invariance models (e.g.~\cite{Foldiak91,wiskott2002slow}) by encouraging the neurons to fire invariantly while transformations are performed in their input stimuli.
However, a model that explains the development of conceptualization from both streams and results in an explicit representation of conceptualization of both streams while the visual stimuli is presented is still missing in the literature. }
This conceptualization should be able to encode the same category for information flows in both ventral and dorsal streams  like `object files' in the visual understanding (\cite{fields2011trajectory}) so that they could be discriminated in different contexts during language development.

On the other hand, this conceptualized representation that is distributed in two visual streams is also able to predict the tendency of appearance of an action-oriented object in the visual field, which causes some sensorimotor phenomena such as object permanence (\cite{tomasello1986object}) showing the infants' attention usually is driven by the object's features and movements.
For instance, when infants are observing the movement of the object, recording showed an increase of the looking times when the visual information after occlusion is violated in either surface features or location (\cite{mareschal2003and}). Also the words and sounds play a top-down role in the early infants' visual attention (\cite{sloutsky2008role}). This could hint at the different development stages of the ventral and dorsal streams and their effect on the conceptualized prediction mechanism in the infant's consciousness.{ {Accordingly, the model we propose about the conceptualized visual information should also be able to explain the emergence of a predictive function in the sensorimotor system, e.g. the ventral stream attempts to track the object and the dorsal stream processes and predicts the object's spatial location, when the sensorimotor system is involved in an object interaction. } We have been aware of that this build-in predictive function in a forward sensorimotor system is essential: neuroimaging research has revealed the existence of internal forward models in the parietal lobe and the cerebellum that predict sensory consequences from efference copies of motor commands (\cite{kawato2003internal}) and supports fast motor reactions (e.g.~\cite{hollerbach1982computers}). 
Since the probable position and the movement pattern of the action should be predicted on a short time scale, sensory feedback produced by a forward model with negligible delay is necessary in this sensorimotor loop. 
Particularly, the predictive sensorimotor model we propose is 
suitable to work as one of the building modules that takes into account the predictive object movement in a forward sensorimotor system to deal with object interaction from visual stimuli input as Fig.~\ref{fig:forward-model} shows. This system is similar to \cite{wolpert1995internal}'s sensorimotor integration, but it includes an additional sensory estimator (the lower brown block) which takes into account the visual stimuli from the object so that it is able to predict the dynamics of both the end-effector (which is accomplished by the upper brown block) and the sensory input of the object. This object-predictive module is essential in a sensorimotor system to generate sensorimotor actions like tracking and avoiding when dealing with fast-moving objects, e.g. in ball sports.
{We also assert that the additional inclusion of forward models in the visual perception of the objects can explain some predictive developmental sensorimotor phenomena, such as object permanence.

\begin{figure}[ht]
\centering
\includegraphics[width=0.5\linewidth]{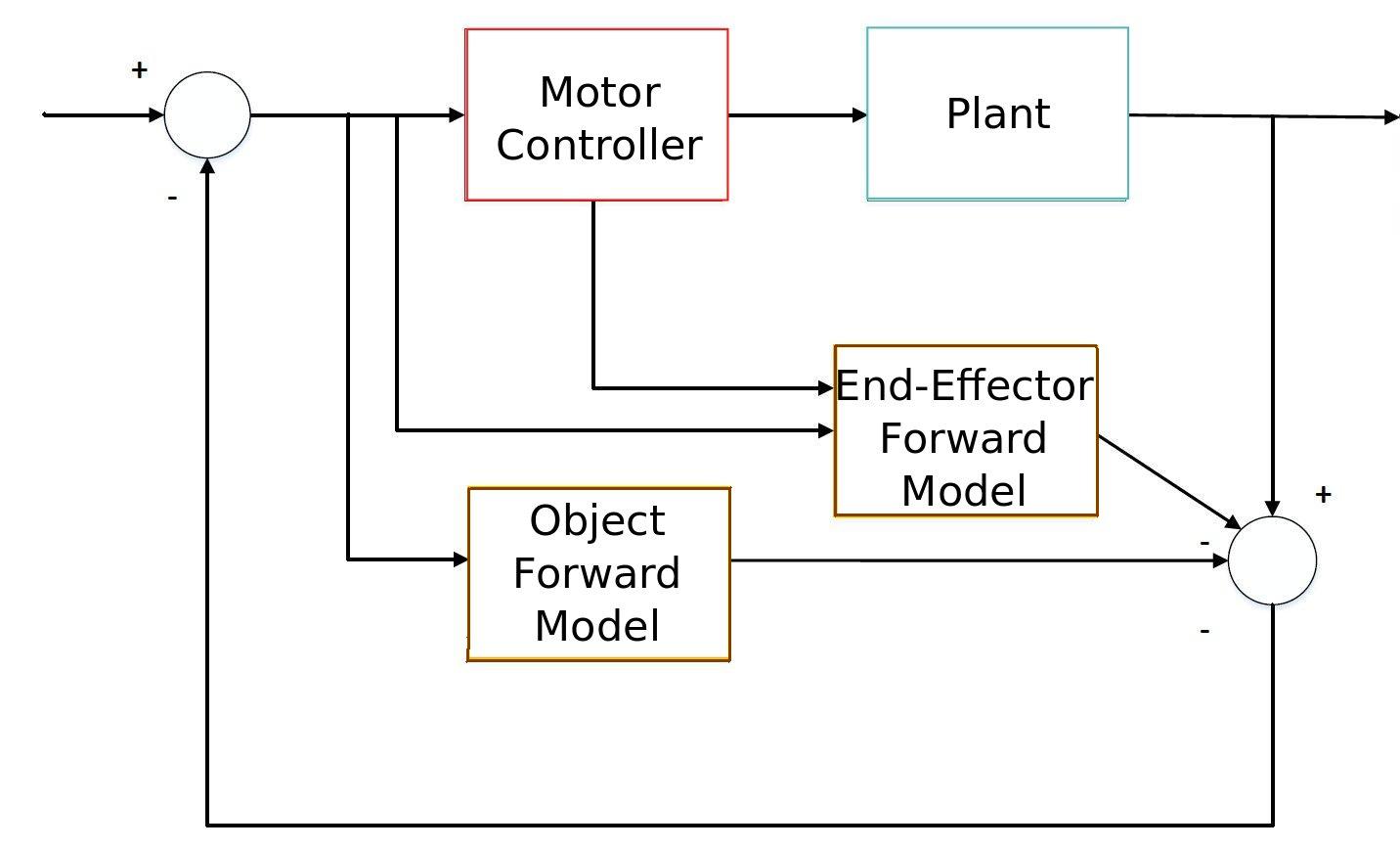}
\caption{Diagram of sensorimotor integration with the object interaction. The lower forward model predicts the object movement, while the upper forward model extracts the end-effector movement from sensory information in order to accomplish a certain task (e.g. object interaction).}
\label{fig:forward-model}
\end{figure}


In summary, we propose a model that establishes links between the development of ventral/dorsal visual streams and the emergence of the conceptualization in visual streams, which further leads to the predictive function of a sensorimotor system. To validate this proof-of-concept model, we also conducted experiments in a simplified robotics scenario. Two NAO robots were employed in the experiments: one of them was used as a `presenter' and moved its arm along pre-programmed trajectories as motion primitives. A ball was attached at the end of the arm so that another robot could obtain the movement by tracking the ball. Our neural network was trained and run on the other NAO, which was called the `observer'. 
In this way, the observer robot perceived the object movement from its vision passively, so that its network took the object's visual features and the movements into account. 
Though we could also use one robot and a human presenter to run the same tasks, we used two identical robots, due to the following reasons: 1. the object movement trajectories can be done by a pre-programmed machinery so that the types and parameters of it can be adjusted; 2. the use of two identical robots allows to interchange the roles of the presenter and observer in an easier manner. 
{As other humanoid robots, a sensorimotor cycle that is composed of cameras and motors also exists in NAO robots. 
Although its physical configurations and parameters of sensory and motor systems are different from those in human beings' or other biological systems, 
our model only handles the pre-processed information extracted from visual stimuli. Therefore it is sufficient to serve as a neural model that is running in a robot CPU to explain the language development in the cortical areas. }  

\begin{figure}[ht]
\centering
\includegraphics[width=0.6\linewidth]{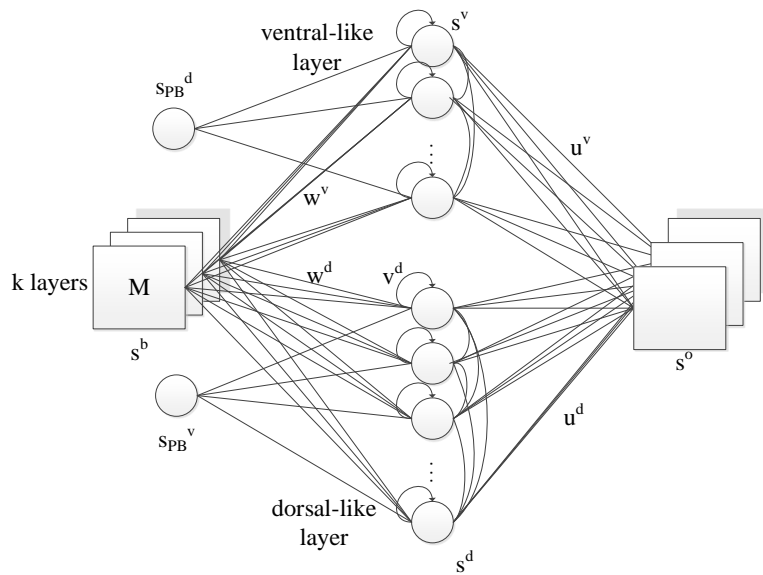}
\caption{The RNNPB-horizontal network architecture, where $k$ layers represent $k$ different types of features. Size of $M$ indicates the transitional information of the object.}
\label{fig:HP-rnn}
\end{figure}

\section{Material and Methods}

\subsection{Network Model}

A similar forward model exhibiting sensory prediction for visual object perception has been proposed in our recently published work (\cite{zhong2013predictive}) where we suggested an RNN implementation of the sensory forward model. Together with a CACLA trained multi-layer network as a controller model, the forward model embodied in a robot receiving visual landmark percepts enabled a smooth and robust robot behavior. 
However, one drawback of this work was its inability to store multiple sets of spatial-temporal input-output mappings, i.e. the learning did not converge if there appeared several spatial-temporal mapping sequences in the training. Consequently, a simple RNN network was not able to predict different sensory percepts for different reward-driven tasks. 
Another problem was that it assumed only one visual feature appeared in the robot's visual field, and that was the only visual cue it could learn during development.  
To solve the first problem, we further augment the RNN with parametric bias (PB) units. They are connected like ordinary biases, but the internal values are also updated through back-propagation. Comparing to the generic RNN, the additional PB units in this network act as bifurcation parameters for the non-linear dynamics. According to~\cite{cuijpers2009generalisation}, a trained RNNPB can successfully retrieve and recognize different types of pre-learned, non-linear oscillation dynamics. Thus, this bifurcation function can be regarded as an expansion of the storage capability of working-memory within the sensory system. Furthermore, it adds the generalization ability of the PB units, in terms of recognizing and generating non-linear dynamics. To tackle the second problem, in order to realize sensorimotor prediction behaviors such as object permanence, the model should be able to learn objects' features and object movements separately in the ventral and dorsal visual streams, as we have shown in~\cite{zhong2012learning}. 

Merging these two ideas, in the context of sensorimotor integration on object interaction, the PB units can be considered as a small set of high-level conceptualized units that describe various types of non-linear dynamics of visual percepts, such as features and movements. This representation is more related to the `natural prototypes' from visual perception, for instance, than a specific language representation (\cite{rosch1973natural}). 

 The development of PB units can also be seen as the pre-symbolic communication that emerges during sensorimotor learning.  The conceptualization, on the other hand, could also result in the prediction of future visual percepts of moving objects in sensorimotor integration.

In this model (Fig.~\ref{fig:HP-rnn}),  we propose a three-layer, horizontal product Elman network with PB units. Similar to the original RNNPB model, the network is capable of being executed under three running modes, according to the pre-known conditions of inputs and outputs: learning, recognition and prediction. In learning mode, the representation of object features and movements are first encoded in the weights of both streams, while the bifurcation parameters with a smaller number of dimensions are encoded in the PB units. This is consistent with the emergence of the conceptualization at the sensorimotor stage of infant development. 

Apart from the PB units, another novelty in the network is that the visual object information is encoded in two neural streams and is further conceptualized in PB units. 
Two streams share the same set of input neurons, where the coordinates of the object in the visual field are used as identities of the perceived images. 
The appearance of values in different layers represents different visual features: in our experiment, the color of the object detected by the yellow filter appears in the first layer whereas the color detected by the green filter appears in the second layer; the other layer remains zero. For instance, the input $((0,0),(x,y))$ represents a green object at $(x, y)$ coordinates in the visual field. 
The hidden layer contains two independent sets of units representing dorsal-like `$d$' and ventral-like `$v$' neurons respectively. These two sets of neurons are inspired by the functional properties of dorsal and ventral streams: (i) fast responding dorsal-like units predict object position and hence encode movements; (ii) slow responding ventral-like units represent object features. The recurrent connection in the hidden layers also helps to predict movements in layer $d$ and to maintain a persistent representation of an object's feature in layer $v$.  
The horizontal product brings both pathways together again in the output layer with one-step ahead predictions. Let us denote the output layer's input from layer $d$ and layer $v$ as $x^d$ and $x^v$, respectively. The network output $s^o$ is obtained via the horizontal product as 

\begin{equation}
\label{eq: horizontal}
s^o = x^d \odot x^v
\end{equation}
where $\odot$ indicates element-wise multiplication, so each pixel is defined by the product of two independent parts, i.e. for output unit $k$ it is $s^o_k = x^d_k \cdot x^v_k$. 


\subsection{Neural Dynamics}
We use  $s^b(t)$ to represent the activation and $PB^{d/v}(t)$ to represent the activation of the dorsal/ventral PB units at the time-step $t$. In some of the following equations, the time-index $t$ is omitted if all activations are from the same time-step. The inputs to the hidden units $y_j^v$ in the ventral stream and $y_j^d$ in the dorsal stream are defined as
\begin{eqnarray}
\label{eq:input}
y^d_l(t) &=& \displaystyle\sum\limits_i s^b_i(t) w^{d}_{li} + \displaystyle\sum\limits_{l'} s^d_l(t-1) v^d_{ll'} + \displaystyle\sum\limits_{n_2} PB_{n_2}^v(t) \bar{w}^{d}_{ln_2} \\
y^v_j(t) &=& \displaystyle\sum\limits_i s^b_i(t) w^{v}_{ji} + \displaystyle\sum\limits_{j'} s^v_j(t-1) v^v_{jj'} + \displaystyle\sum\limits_{n_1} PB_{n_1}^d(t) \bar{w}^{v}_{jn_1}  
\end{eqnarray}
\noindent where  $w^d_{li}$, $w^v_{ji}$ represent the weighting matrices between dorsal/ventral layers and the input layer, $\bar{w}_{li}^d$, $\bar{w}_{ji}^v$  represent the weighting matrices between PB units and the two hidden layers, and $v^d_{ll'}$ and $v^v_{jj'}$ indicate the recurrent weighting matrices within the hidden layers. 

The transfer functions in both hidden layers and the PB units all employ the sigmoid function recommended by~\cite{lecun1998efficient}, 
\begin{eqnarray}
s^{d/v}_{l/j} = 1.7159 \cdot tanh (\frac{2}{3}  y^{d/v}_{l/j}) \label{eq:hidden} \\
\label{eq: PB-trans}
PB^{d/v}_{n_1/n_2} = 1.7159 \cdot tanh (\frac{2}{3}  \rho^{d/v}_{n_1/n_2})
\end{eqnarray}
\noindent where $\rho^{d/v}$ represent the internal values of the PB units.

The terms of the horizontal products of both pathways can be presented as follows:
\begin{eqnarray}
\label{eq:hiddenoutputbetween}
x^v_k = \displaystyle\sum\limits_j s^v_j  u^{v}_{kj};  \; \; \; x^d_k = \displaystyle\sum\limits_l s^d_l  u^{d}_{kl}
\end{eqnarray}
The output of the two streams composes a horizontal product for the network output as we defined in Eq.~\ref{eq: horizontal}.

\subsubsection{Learning mode}
The training progress is basically determined by the cost function:
\begin{equation}
\label{eq:energy}
    C = \frac{1}{2} \sum^T_t \sum_k^N (s_{k}^b(t+1) - s_{k}^o(t))^2
\end{equation}
\noindent where $s_{i}^b(t+1)$ is the one-step ahead input (as well as the desired output), $s_{k}^o(t)$ is the current output, $T$ is the total number of available time-step samples in a complete sensorimotor sequence and $N$ is the number of output nodes, which is equal to the number of input nodes. Following gradient descent, each weight update in the network is proportional to the negative gradient of the cost with respect to the specific weight $w$ that will be updated:
\begin{equation}
\Delta w_{ij} = -\eta_{ij} \frac {\partial C}{\partial w_{ij}}
\end{equation}

\noindent where $\eta_{ij}$ is the adaptive learning rate of the weights between neuron $i$ and $j$, which is adjusted in every epoch (\cite{kleesiek2013action}). To determine whether the learning rate has to be increased or decreased, we compute the changes of the weight $w_{i,j}$ in consecutive epochs:
\begin{equation}
\sigma_{i,j} = \frac{\partial C}{\partial w_{i,j}}(e-1)
\frac{\partial C}{\partial w_{i,j}}(e)
\end{equation}

The update of the learning rate is 
\[
\eta_{i,j}(e) = \begin{dcases*}
min(\eta_{i,j}(e-1) \cdot \xi^+, \eta_{max}) & {if $ \sigma_{i,j} > 0$}, \\
max(\eta_{i,j}(e-1) \cdot \xi^-, \eta_{min}) & {if $ \sigma_{i,j} < 0$}, \\
\eta_{i,j}(e-1) & else. \end{dcases*} 
\]
\noindent where $\xi^+ > 1 $ and $\xi ^- < 1$ represent the increasing/decreasing rate of the adaptive learning rates, with $\eta_{min}$ and $\eta_{max}$ as lower and upper bounds, respectively. Thus, the learning rate of a particular weight increases by $\xi^+$ to speed up the learning when the changes of that weight from two consecutive epochs have the same sign, and vice versa. 

Besides the usual weight update according to back-propagation through time, the accumulated error over the whole time-series also contributes to the update of the PB units. The update for the $i$-th unit in the PB vector for a time-series of length $T$ is defined as:

\begin{equation}
\label{eq: PB-error}
\rho_i (e+1) = \rho_i (e) + \gamma_i \sum_{t=1}^T \delta_{i,j}^{PB}\\
\end{equation}
where $\delta^{PB}$ is the error back-propagated to the PB units, $e$ is $e$th time-step in the whole time-series (e.g. epoch), $\gamma_i$ is PB units' adaptive updating rate which is proportional to the absolute mean value of the back-propagation error at the $i$-th PB node over the complete time-series of length $T$:

\begin{equation}
\gamma_i \propto \frac{1}{T} \norm{\sum_{t=1}^T \delta_{i,j}^{PB}}
\end{equation}

The reason for applying the adaptive technique is that it was realized that the PB units converge with difficulty. Usually a smaller learning rate is used in the generic version of RNNPB to ensure the convergence of the network. However, this results in a trade-off in convergence speed. The adaptive learning rate is an efficient technique to overcome this trade-off (\cite{kleesiek2013action}).

\subsubsection{Recognition mode}
The recognition mode is executed with a similar information flow as the learning mode: given a set of the spatio-temporal sequences, the error between the target and the real output is back-propagated through the network to the PB units. However, the synaptic weights remain constant and only the PB units will be updated, so that the PB units are self-organized as the pre-trained values after certain epochs. Assuming the length of the observed sequence is $a$, the update rule is defined as:   

\begin{equation}
\label{eq:PB_recognition}
\rho_i (e+1) = \rho_i (e) + \gamma \sum_{t=T-a}^T \delta_{i,j}^{PB}
\end{equation}

\noindent where $\delta^{PB}$ is the error back-propagated from a certain sensory information sequence to the PB units and $\gamma$ is the updating rate of PB units in recognition mode, which should be larger than the adaptive rate $\gamma_i$ at the learning mode.

\subsubsection{Prediction mode}
The values of the PB units can also be manually set or obtained from recognition, so that the network can generate the upcoming sequence with one-step prediction. 


\begin{table}[h]
\centering
\begin{tabular}[c]{c|c|c}
\hline
Parameters & Parameter's Descriptions & Value\\
\hline
$\eta_{ventral}$ & Learning Rate in Ventral Stream & $1.0 \times 10^{-5}$ \\
$\eta_{dorsal}$ & Learning Rate in Dorsal Stream & $1.0 \times 10^{-3}$ \\ 
$ \eta_{max} $ & Maximum Value of Learning Rate & $1.0 \times 10^{-1}$ \\
$ \eta_{min} $ & Minimum Value of Learning Rate & $1.0 \times 10^{-7}$ \\
$ M_{\gamma} $ & Proportionality Constant of PB Units Updating Rate & $1.0 \times 10^{-2}$ \\
$n_1 $ & Size of PB Unit 1 & $1$ \\
$n_2 $ & Size of PB Unit 2 & $1$ \\
$n_v$ & Size of Ventral-like Layer & $50$ \\
$n_d$ & Size of Dorsal-like Layer & $50$ \\
$\xi^-$ & Decreasing Rate of Learning Rate & $0.999999$ \\
$\xi^+$ & Increasing Rate of Learning Rate & $1.000001$ \\
\hline
\end{tabular}
\caption{Network parameters}
\label{tab:parameter}
\end{table}

\section{Results}

{In this experiment, as we introduced, we examined this network by implementing it on two NAO robots.} They were placed face-to-face in a rectangle box of $61.5cm \times 19.2cm $  as shown in Fig.~\ref{fig:robots}. These distances were carefully adjusted so that the observer was able to keep track of movement trajectories in its visual field during all experiments {using the images from the lower camera.} {The NAO robot has two cameras. We use the lower one to capture the images because its installation angle is more suitable to track the balls when they are held in the other NAO's hand. }

\begin{figure}[h]
\centering
\includegraphics[width=0.3\linewidth]{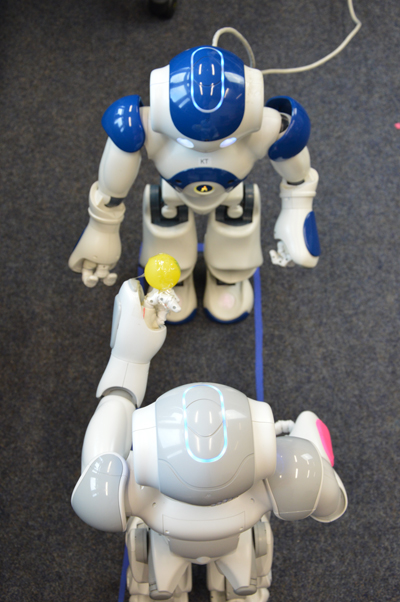} 
\caption{Experimental Scenario: two NAOs are standing face-to-face with in a rectangle box.}
\label{fig:robots}
\end{figure}
Two $3.8cm$-diameter balls with yellow/green color were used for the following experiments. The presenter consecutively held each of the balls to present the object interaction.  
The original image, received from the lower camera of the observer, was pre-processed with thresholding in HSV color-space and the coordinates of its centroid in the image moment were calculated. Here we only considered two different colors as the only feature to be encoded in the ventral stream, as well as two sets of movement trajectories encoded in the dorsal stream. Although we have only tested a few categories of trajectories and features, we believe the results can be extrapolated to multiple categories in future applications.

\subsection{Learning}

The two different trajectories are defined as below,

\noindent The \textit{cosine} curve,
\begin{eqnarray}
\label{eq:cos}
x &=&  12\\
y &=&  8 \cdot (-\frac{t}{2}) + 0.04\\
    z &=&  4 \cdot \cos(2t) + 0.10
\end{eqnarray}

\noindent and the \textit{square} curve,
\begin{eqnarray}
x  &=&   12 \\
y  &=&   \left\{
  \begin{array}{ll}
    0 & t \leq -\frac{3\pi}{4}\\
    \frac{16}{\pi}t + 12 & -\frac{3\pi}{4} < t \leq -\frac{\pi}{4} \\
    8 & -\frac{\pi}{4} < t \leq \frac{\pi}{4}\\
    -\frac{16}{\pi}t + 12 & \frac{\pi}{4} < t \leq \frac{3\pi}{4} \\
    0 & t > \frac{3\pi}{4}
  \end{array} \right.\\
z &=& \left\{
  \begin{array}{ll}
     \frac{16}{\pi}t + 20 & t \leq -\frac{3\pi}{4}\\
     14 & -\frac{3\pi}{4} < t \leq -\frac{\pi}{4} \\
    -\frac{16}{\pi}t + 10 & -\frac{\pi}{4} < t \leq \frac{\pi}{4} \\
    6 & \frac{\pi}{4} < t \leq \frac{3\pi}{4}\\
    \frac{16}{\pi}t - 6 & t > \frac{3\pi}{4}   \\
     \end{array} \right. \label{eq:square}
\end{eqnarray}

\noindent where the 3-dimension tuple $(x,y,z)$ are the coordinates (centimetres) of the ball w.r.t the torso frame of the NAO presenter. $t$ loops between $(-\pi, \pi]$. In each loop, we calculated 
$20$ data points to construct trajectories with $4s$ sleeping time between every two data points. Note that although we have defined the optimal desired trajectories, the arm movement was not ideally identical to the optimal trajectories due to the noisy position control of the end-effector of the robot.  
{On the observer side, the $(x,y)$ coordinates of the color-filtered moment of the ball in the visual field were recorded to form a trajectory with sampling time of $0.2s$. Five trajectories, in the form of tuple $(x,y,z)$ w.r.t the torso frame of the NAO observer were recorded with each color and each curve, so total $20$ trajectories were available for training. }

In each training epoch, these trajectories, in the form of tuples, were fed into the input layer one after another for training, with the tuples of the next time-step serving as a training target. The parameters are listed in Tab.~\ref{tab:parameter}. 
The final PB values were examined after the training was done, and the values were shown in Fig.~\ref{fig:training}. It can be seen that the first PB unit, along with the dorsal stream, was approximately self-organized with the color information, while the second PB unit, along with the ventral stream, was self-organized with the movement information.   
\begin{figure}[h]
\centering
\includegraphics[width=0.7\linewidth]{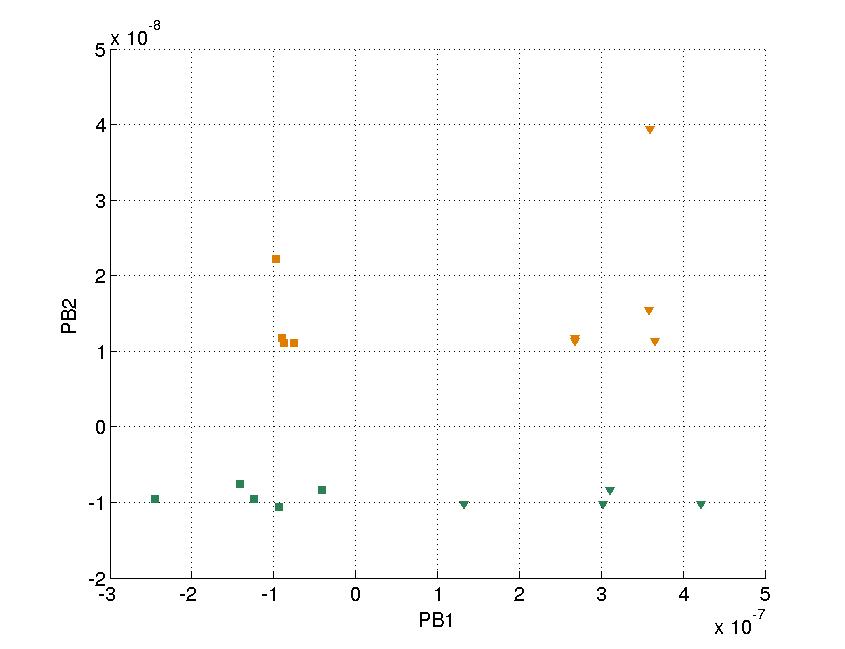} 
\caption{Values of two sets of PB units in the two streams after training. The square markers represent those PB units after the \textit{square} curves training and the triangle markers represent those of the \textit{cosine} curves training. The colors of the markers, yellow and green, represent the colors of the balls used for training.}
\label{fig:training}
\end{figure}

\begin{figure}[h]
\centering
    \subfloat[PB value 1\label{subfig-1:rec1}]{%
      \includegraphics[width=0.45\textwidth]{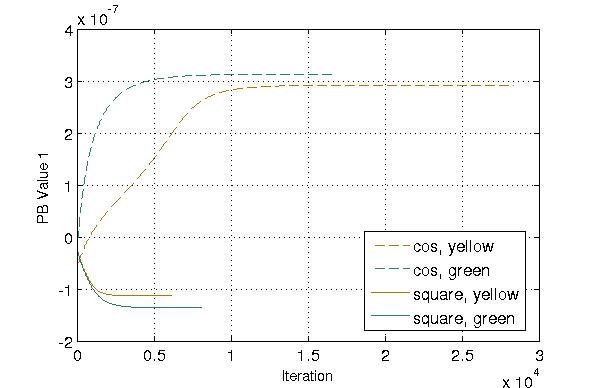}
    } 
     \subfloat[PB value 2\label{subfig-2:rec2}]{%
      \includegraphics[width=0.45\textwidth]{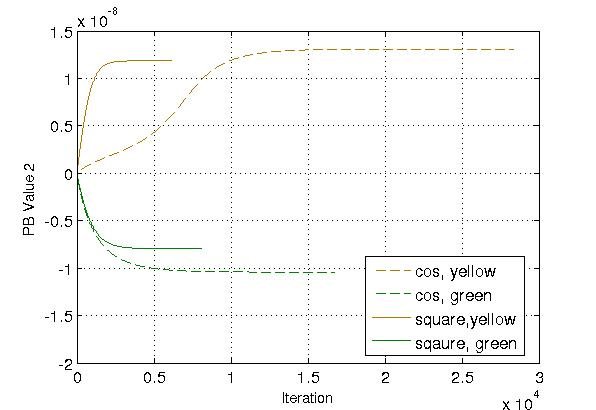}
    }
    \caption{Update of the PB values while executing the recognition mode}
    \label{fig:rec}
  \end{figure}


\subsection{Recognition}
Another four trajectories were presented in the recognition experiment, in which the length of the sliding-window is equal to the length of the whole time-series, i.e. $T = a$ in Eq.~\ref{eq:PB_recognition}. The update of the PB units were shown in Fig.~\ref{fig:rec}. Although we used the complete time-series sequence for the recognition, it should also be possible to use only part of the sequence, e.g. through the sliding-window approach with a smaller number of $a$ to fulfil the real-time requirement in the future.

\begin{figure}[h]
\centering
\subfloat[Cosine curve, yellow ball\label{subfig-3:cos2}]{%
      \includegraphics[width=0.45\textwidth]{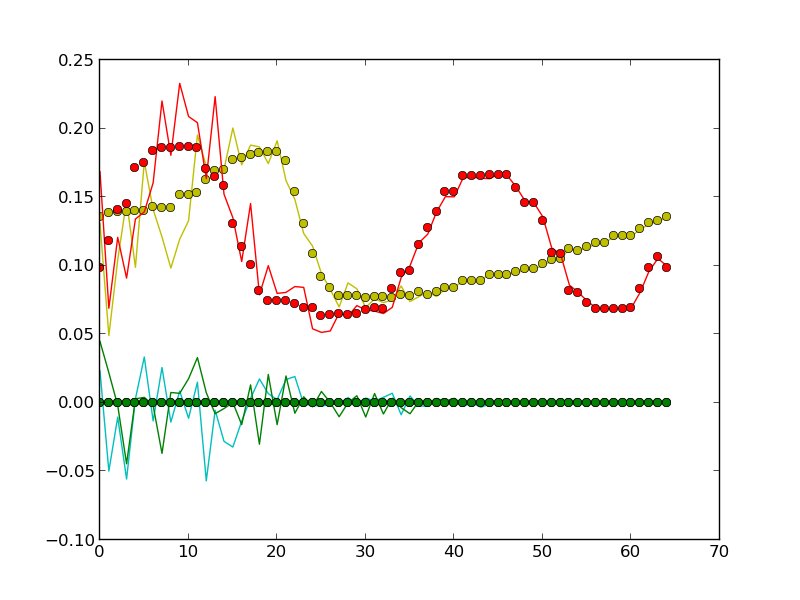}
    }
    \subfloat[Cosine curve, green ball\label{subfig-4:cos2}]{%
      \includegraphics[width=0.45\textwidth]{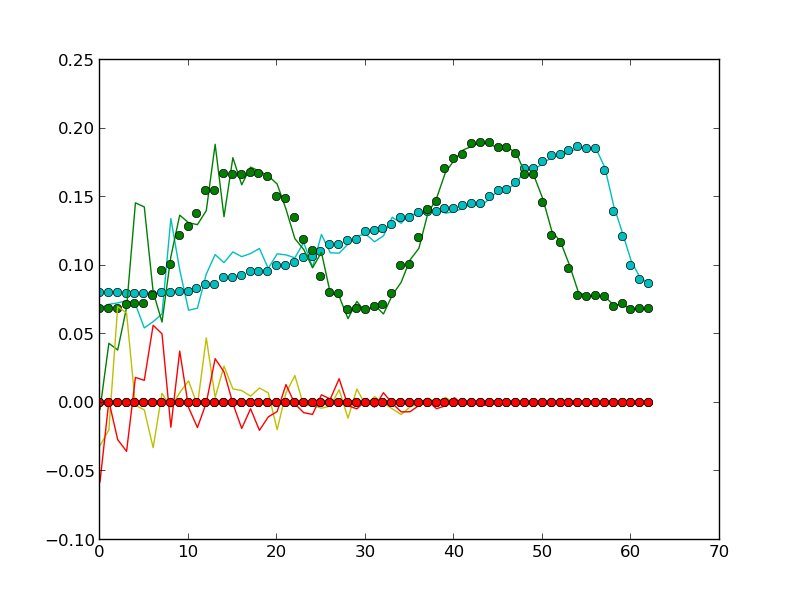}
    }\\
    \subfloat[Square curve,  yellow ball\label{subfig-1:sq1}]{%
      \includegraphics[width=0.45\textwidth]{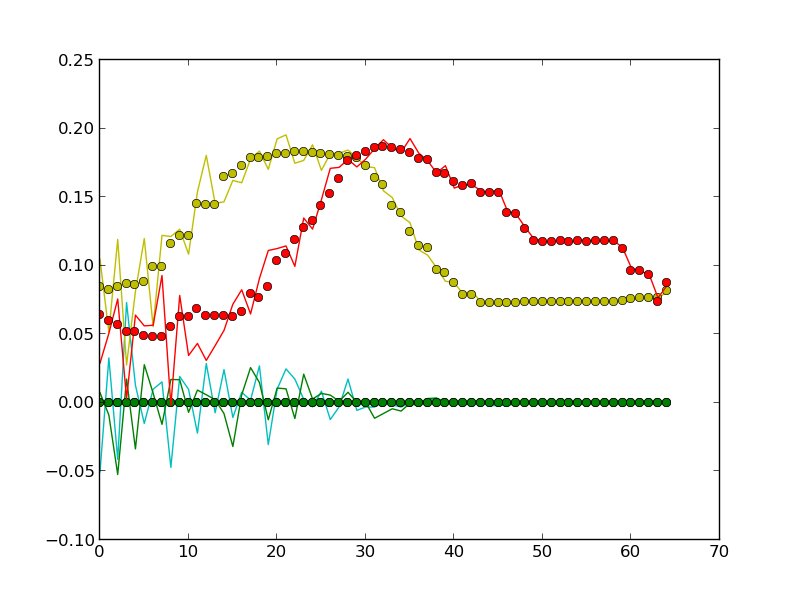}
    }
     \subfloat[Square curve, green ball\label{subfig-2:sq2}]{%
      \includegraphics[width=0.45\textwidth]{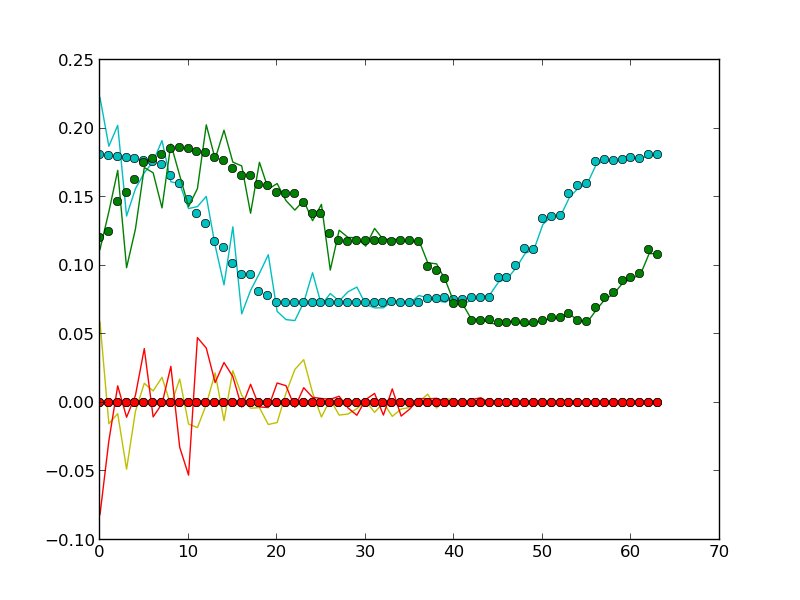}
    }\\
    
    \caption{Generated Values: the dots denote the true values for comparison, curves show the estimated ones. Yellow and red colors represent the values of the two neurons in the first layer (yellow), the colors green and clan represent those in the second layer (green).}
    \label{fig:prediction}
  \end{figure} 

\subsection{Prediction}
In this simulation, the obtained PB units from the previous recognition experiment were used to generate the predicted movements using the prior knowledge of a specific object. Then, the one-step prediction from the output units were again applied to the input at the next time-step, so that the whole time-series corresponding to the object's movements and features were obtained.  Fig.~\ref{fig:prediction} presents the comparisons between the true values (the same as used in recognition) and the predicted ones.


From Fig.~\ref{fig:prediction}, it can be observed that the estimation was biased quite largely to the true value within the first few time-steps, as the RNN needs to accumulate enough input values to access its short-term memory. However, the error became smaller and it kept track of the true value in the following time-steps. Considering that the curves are automatically generated given the PB units and the values at the first time-step, the error between the true values and the estimated ones are acceptable. Moreover, this result show clearly that the conceptualization affects the (predictive) visual perception.

\begin{table}
\centering
\begin{tabular}{c|c|c|c|c}
\hline
Error of Outputs & Unit 1 & Unit 2 & Unit 3 & Unit 4\\
\hline
cosine, yellow & $2.28 \times 10^{-4}$ & $8.09 \times 10^{-5}$ & $7.29 \times 10^{-4}$ & $8.63 \times 10^{-4}$\\
cosine, green  & $8.34 \times 10^{-4}$ & $7.04 \times 10^{-4}$ & $1.50 \times 10^{-4}$ & $2.01 \times 10^{-4}$ \\ 
square, yellow & $3.91 \times 10^{-4}$ & $9.64 \times 10^{-5}$ & $1.74 \times 10^{-3}$ & $3.23 \times 10^{-4}$ \\
square, green  &  $1.40 \times 10^{-3}$ & $3.27 \times 10^{-4}$ & $3.54 \times 10^{-4}$ & $2.60 \times 10^{-4}$ \\
\hline
\end{tabular}
\caption{Prediction error}
\label{tab:predict}
\end{table}

\subsection{Generalization in Recognition}
To testify whether our new computational model has the generalization ability as~\cite{cuijpers2009generalisation} proposed, we recorded another set of sequences of a circle trajectory. The trajectory is defined as:

\begin{eqnarray}
\label{eq:circle}
x &=&  12\\
y &=&  4 \cdot \sin (2t) + 0.04\\
    z &=&  4 \cdot \cos(2t) + 0.10
\end{eqnarray}

The yellow and green balls were still used. We ran the recognition experiment again with the weight previously trained. The update of the PB units were shown in Fig.\ref{fig:rec-generalisation}. Comparing to Fig.~\ref{fig:training}, we can observe that the positive and negative signs of PB values are similar as the square trajectory. This is probably because the visual perception of circle and square movements have more similarities than those between circle and cosine movements. 
\begin{figure}[h]
\centering
    \subfloat[PB value 1\label{subfig-1:rec1}]{%
      \includegraphics[width=0.45\textwidth]{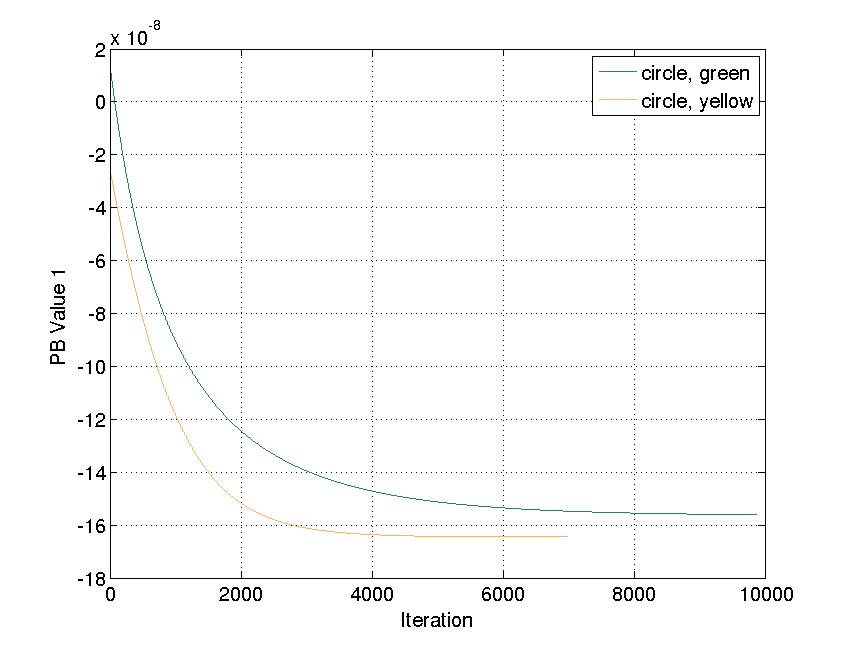}
    } 
     \subfloat[PB value 2\label{subfig-2:rec2}]{%
      \includegraphics[width=0.45\textwidth]{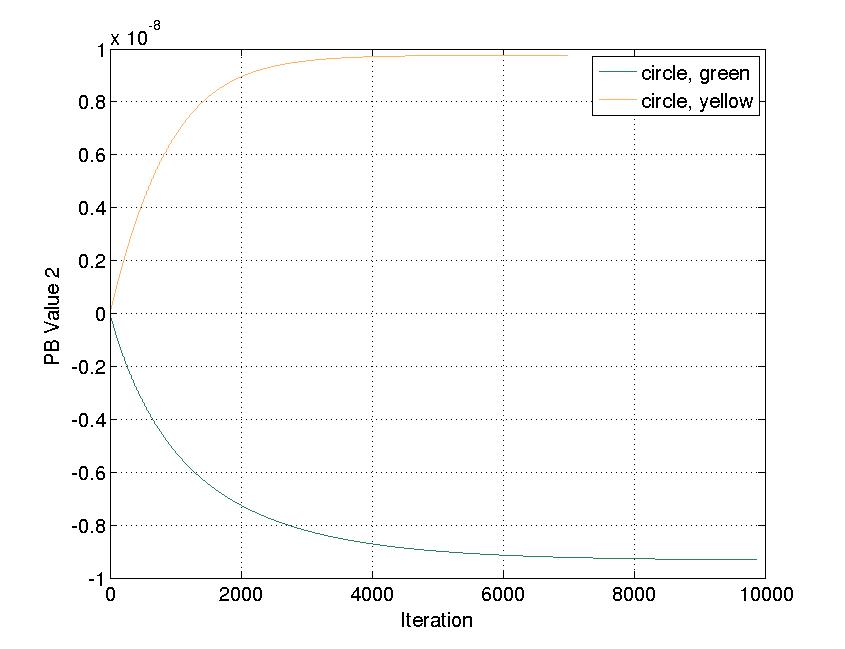}
    }
    \caption{Update of the PB values while executing the recognition mode with an untrained feature (circle)}
    \label{fig:rec-generalisation}
  \end{figure}

\subsection{PB representation with different speeds}
We further generated $20$ trajectories with the same data functions (Eqs. \ref{eq:cos} - \ref{eq:square}) but with a slower sampling time. In other words, the movement of the balls seemed to be faster with robot's observation. The final PB values after training were shown in Fig.~\ref{fig:training-fast}. 

It can be seen that generally the PB values were smaller comparing to Fig.~\ref{fig:training}, which was probably because there was less error being propagated during training. Moreover, the corresponding PB values corresponding to colors (green and yellow) and movements (cosine and square) were interchanged within the same PB unit (i.e. along the same axis) due to the difference of random initial parameters of the network. But the PB unit along with the dorsal stream still encoded color information, while the PB unit along with ventral stream encoded movement information. The network was still able to show properties of spatio-temporal sequences data in the PB units' representation.   

\begin{figure}[h]
\centering
\includegraphics[width=0.7\linewidth]{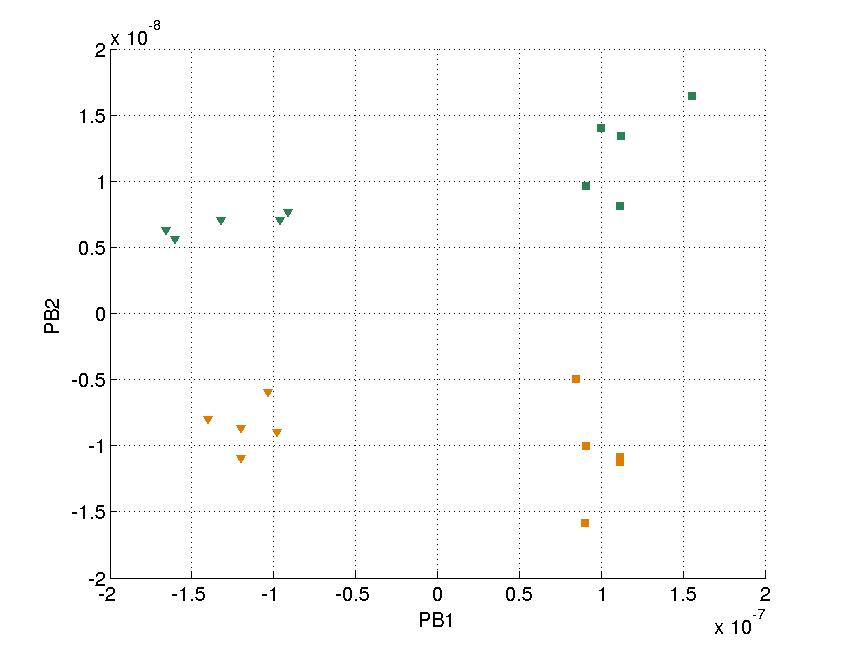} 
\caption{Values of two sets of PB units in the two streams after training with faster speed. The representation of the markers is the same as Fig.~\ref{fig:training}}
\label{fig:training-fast}
\end{figure}

\section{Discussion}

\subsection{Neural Dynamics}
An advancement of the HP-RNN model is that it can learn and encode the `what' and `where' information separately in two streams (more specifically, in two hidden layers). Both streams are connected through horizontal products, which means fewer connections than full multiplication (as the conventional bilinear model) (\cite{zhong2012learning}). In this paper, we further augmented the HP-RNN with the PB units.  One set of units, connecting to one visual stream, reflects the dynamics of sequences in the other stream. This is an interesting result since it shows the neural dynamics in the hybrid combination of the RNNPB units and the horizontal product model. 
Taking the dorsal-like hidden layer for example, the error of the attached PB units is

\begin{eqnarray}
\delta_{n_2}^{PB} &=&  \sum_{l} \delta_l^d \cdot f'(\rho^v_{n_2}) \cdot \bar{w}^{d}_{ln_2} \\
&=& \sum_{l} \left[\sum_{k} (s_{k}^b-s_{k}^o) g'(s_k^o) u_{kl}^d f'(s^v_{l})  \right] f'(\rho^v_{n_2}) \cdot \bar{w}^{d}_{ln_2} 
\end{eqnarray}

\noindent where  $g'(\cdot)$ and $f'(\cdot)$ are the derivatives of the linear and sigmoid transfer functions. Since we have the linear output, according to the definition of the horizontal product, the equation becomes,
\begin{equation}
\delta_{n_2}^{PB} = \sum_{l} \left[\sum_{k} (s_{k}^b-s_{k}^o) \odot x_k^v \cdot u_{kl}^d f'(s^d_{l}) \right] f'(\rho^v_{n_2}) \cdot \bar{w}^{d}_{ln_2} 
\end{equation}

The update of the internal values of the PB units becomes
\begin{eqnarray}
\rho_{n_2}^v (e+1) &=& \rho_i^v (e) + \gamma \sum_{t=T-a}^T \delta_{n_2}^{PB}(t) \\
&=& \rho_i^v (e) + \gamma \sum_{t=T-a}^T \Big\{\sum_{l}  \big[ \sum_{k} (s_{k}^b(t)-s_{k}^o(t)) \odot x_k^v(t) \cdot u_{kl}^d(t) f'(s^d_{l})\big]f'(\rho_{n_2}^v(t))\cdot \bar{w}^{d}_{ln_2}(t)  \Big \} \nonumber \\
\end{eqnarray}

\noindent where the $x_k^v(t) $ term refers to the contribution of the weighted summation from the ventral-like layer at time $t$. Note that the term $f'(\rho_{n_2}^v(t))$ is actually constant within one epoch and it is only updated after each epoch with a relatively small updating rate. Therefore, from the experimental perspective, given the same object movement but different object features, the difference of the PB values mostly reflects the dynamic changes in the hidden layer of the ventral stream. The same holds for the PB units attached to the ventral-like layer. This brief analysis shows the PB units for one modularity in RNNPB networks with horizontal product connections, effectively accumulating the non-linear dynamics of other modularities. 

\subsection{Conceptualization in visual perception}
The visual conceptualization and perception are intertwined processes. As experiments from~\cite{schyns1999dr} show,  when the visual observation is not clear, the brain automatically extrapolates the visual percept and updates the categorization labels on various levels according to what has been gained from the visual field. 
On the other hand, this conceptualization also affects the immediate visual perception in a top-down predictive manner. 
For instance, the identity conceptualization of a human face predictively spreads conceptualizations in other levels (e.g. face emotion). This top-down process propagates from object identity to other local conceptualizations, such as object affordance, motion, edge detection and other processes at the early stages of visual processing. This can be tested by classic illusions, such as `the goblet illusion', where perception depends largely on top-down knowledge derived from past experiences rather than direct observation. This kind of illusion may be explained by the error in the first few time steps of the prediction experiment of our model.
Therefore, our model to some extent also demonstrates the integrated process between the conceptualization and the spatio-temporal visual perception. This top-down predictive perception may also arouse other visual based predictive behaviors such as object permanence.

Particularly, the PB units act as a high-level conceptualization representation, which is continuously updated with the partial sensory information perceived in a short-time scale. 
The prediction process of the RNNPB is assisted by the conceptualized PB units of visual perception, which is identical to the integration conceptualization and (predictive) visual perception.
This is the reason why PB units were not processed as a binary representation, as \cite{ogata2007human} did for human-robot-interaction; the original values of PB units are more accurate in generating the prediction of the next time-step and performing generalization tasks. 
As we mentioned, this model is merely a proof-of-concept model that bridges conceptualized visual streams and sensorimotor prediction. 
For more complex tasks, besides expanding of the network size as we mentioned, more complex networks that are capable of extracting and predicting higher-level spatio-temporal structures (e.g. predictive recurrent networks owning large learning capacity by Tani and colleagues:  \cite{yamashita2008emergence, murata2013learning}) can be also applied. It should be interesting to further investigate the functional modularity representation of these network models when they are interconnected with horizontal product too.

{Furthermore, the neuroscience basis that supports this paper, in the context of the mirror neuron system based on object-oriented-actions (grasping), can be stated as the `data-driven' models such as MNS (\cite{oztop2002schema}) and MNS2 (\cite{bonaiuto2007extending,bonaiuto2010extending}), although the main hypothesis in our model is not taken from the mirror neuron system theory. In the MNS review paper by \cite{oztop2006mirror}, the action generation mode of the RNNPB model was considered to be excessive as there has no evidence yet to show that the mirror neuron system participates in action generation. However, in our model the generation mode has a key role of conceptualized PB units in the sensorimotor integration of object interaction. Nevertheless, the similar network architecture (RNNPB) used in modeling mirror neurons (\cite{tani2004self}) and our pre-symbolic sensorimotor integration models may imply a close relationship between language (pre-symbolic) development, object-oriented actions, and the mirror neuron theory.}
  
\section{Conclusion}
{In this paper a recurrent network architecture integrating the RNNPB model and the horizontal product model has been presented, which sheds light on the feasibility of linking the conceptualization of ventral/dorsal visual streams, the emergence of pre-symbol communication, and the predictive sensorimotor system. 

Based on the horizontal product model, here the information in the dorsal and ventral streams is separately encoded in two network streams and the predictions of both streams are brought together via the horizontal product while the PB units act as a conceptualization of both streams. These PB units allow for storing multiple sensory sequences. After training, the network is able to recognize the pre-learned conceptualized information and to predict the up-coming visual perception. The network also shows robustness and generalization abilities. 
Therefore, our approach offers preliminary concepts for a similar development of conceptualized language in pre-symbolic communication and further in infants' sensorimotor-stage learning. } 
 
\section*{Acknowledgement}
The authors thank Sven Magg, Cornelius Weber, Katja K\"{o}sters as well as reviewers (Matthew Schlesinger, Stefano Nolfi) for improvement of the paper, Erik Strahl for technical support in Hamburg and Torbjorn Dahl for the generous allowance to use the NAOs in Plymouth.

\paragraph{Funding\textcolon} 
This research has been partly supported by the EU projects RobotDoC under 235065 ROBOT-DOC from the 7th Framework Programme (FP7), Marie Curie Action ITN, and KSERA funded from FP7 for Research and Technological Development under grant agreement n\textdegree 2010-248085, POETICON++ under grant agreement 288382 and UK EPSRC project BABEL.

\bibliographystyle{plainnat}

\bibliography{prediction}

\end{document}